\documentclass[10pt,twocolumn,letterpaper]{article}

\usepackage{cvpr}              

\usepackage[pagebackref,breaklinks,colorlinks]{hyperref}

\usepackage{graphicx}
\usepackage{caption}
\usepackage{subcaption} 
\usepackage{amsmath}
\usepackage{amsfonts}
\usepackage{algorithm}
\usepackage{algorithmic}
\usepackage{subcaption}
\usepackage{lipsum}
\usepackage{graphicx}
\usepackage{mathrsfs}
\usepackage{multirow}
\usepackage{bm}
\usepackage{amsthm,amsmath,amssymb}
\usepackage{dsfont}
\usepackage{booktabs}
\usepackage{xcolor}
\usepackage{lipsum}
\usepackage{natbib}
\usepackage{hyperref}

\title{ A brief introduction to a framework named Multilevel Guidance-Exploration Network }

\author{Guoqing Yang$^{1}$\qquad Zhiming Luo$^{1}$ \qquad Jianzhe Gao$^{1}$  \qquad Yingxin Lai$^{1}$ \qquad \\Kun Yang$^{1}$\qquad Yifan He$^{2}$ \qquad Shaozi Li$^{1}$\\
$^1$ Xiamen University,Xiamen, China;$^{2}$Reconova Technologies,  China}

\begin{document}
\maketitle

\begin{abstract}
Human behavior anomaly detection aims to identify unusual human actions, playing a crucial role in intelligent surveillance and other areas. The current mainstream methods still adopt reconstruction or future frame prediction techniques.  However, reconstructing or predicting low-level pixel features easily enables the network to achieve overly strong generalization ability, allowing anomalies to be reconstructed or predicted as effectively as normal data. Different from their methods, inspired by the Student-Teacher Network, we propose a novel framework called the Multilevel Guidance-Exploration Network(MGENet), which detects anomalies through the difference in high-level representation between the Guidance and Exploration network. Specifically, we first utilize the pre-trained Normalizing Flow that takes skeletal keypoints as input to guide an RGB encoder, which takes unmasked RGB frames as input, to explore motion latent features. Then, the RGB encoder guides the mask encoder, which takes masked RGB frames as input, to explore the latent appearance feature. Additionally, we design a Behavior-Scene Matching Module(BSMM) to detect scene-related behavioral anomalies. Extensive experiments demonstrate that our proposed method achieves state-of-the-art performance on ShanghaiTech and UBnormal datasets. 

\end{abstract}

\section{Introduction}
Human behavior anomaly detection aims to temporally or spatially localize the abnormal actions of the person within a video. It plays a significant role in enhancing public security~\cite{a1,hsc}. Detecting such anomalies presents a challenge due to the infrequent occurrence and the various types of abnormal events~\cite{GCL}. As a result, most typical methods~\cite{memory,STG-NF,jigsaw,aaai21-Consistency,keyframe,liu2018future}, employ unsupervised learning approaches using only normal data for training. In these approaches, including our method, behaviors that the model identifies as outliers are considered anomalies.

\begin{figure}[t]
    \setlength{\abovecaptionskip}{0.cm}
    \begin{center}
        \includegraphics[width=1.05\linewidth]{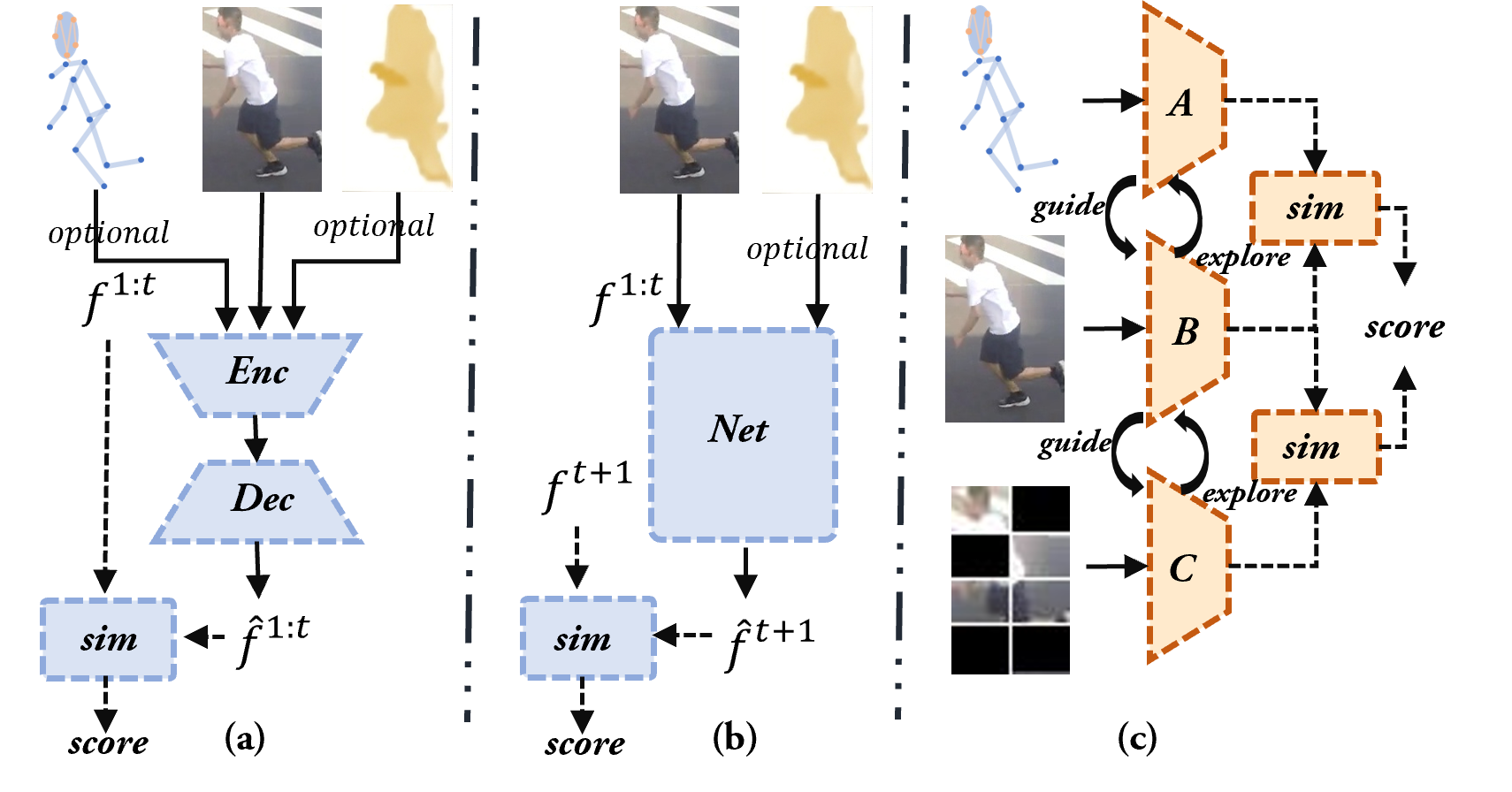}
    \end{center}
    \caption{Comparison of different methods using various features.  (a) Reconstruction-based method, using the autoencoder to reconstruct the previous $T$ frames $f^{1:t}$. (b) Prediction-based method,  predicting the $t+1$ frame $f^{t+1}$ from the prior $T$ frames. Both of them detect anomalies based on reconstruction or prediction errors.  (c) Our Multilevel Guidance-Exploration framework, includes two similar levels. For instance, in the $1$-st level, Encoder-B learns another type of feature under the guidance of a pre-trained network (Encoder-A), detecting anomalies based on the similarity of latent output features.}
    \label{fig:introduction}
\end{figure}

Many unsupervised methods often use reconstruction or future frame prediction methods combined with various features to detect human behavior anomalies. The reconstruction-based framework~\cite{memory,sspcab,wang2022making,ssmtl,liu2021hybrid}, illustrated in Figure~\ref{fig:introduction}(a), utilizes autoencoders trained on normal data, detecting anomalies based on elevated reconstruction errors. ~\cite{wang2022making} propose a new autoencoder model, named Spatio-Temporal Auto-Trans-Encoder, to enhance consecutive frame reconstruction.  As depicted in Figure ~\ref{fig:introduction}(b), the prediction-based methods~\cite{liu2018future,Consistency,aaai21-Consistency,keyframe,ssmtl++,liu2021hybrid} typically predict pixel-level features for the next frame using previous frames. ~\cite{aaai21-Consistency} propose an Appearance-Motion Memory Consistency Network based on autoencoders, explicitly considering the endogenous consistency semantics between optical flow features and RGB appearance features during the prediction process. Additionally, Liu et al.~\cite{liu2021hybrid} use a hybrid strategy by initially reconstructing optical flow features with a reconstruction autoencoder and then jointly predicting the next frame with previous frames.

However, reconstructing or predicting pixel-level features at a low level can result in the network having overly strong generalization~\cite{jigsaw,keyframe,aaai21-Consistency,memory}, where some anomalous samples can be reconstructed or predicted as effectively as normal samples. This phenomenon poses a challenge in distinguishing between normal and anomalous instances. Additionally, these approaches~\cite{jigsaw,STG-NF,liu2021hybrid,ssmtl,ssmtl++,Consistency,georgescu2021background} ignore scene context. They focus solely on the behavior of individuals without considering their interaction with the surrounding scene. For example, lying on a zebra-crossing road is more likely to be considered anomalous compared to the same posture on a beach setting.

Different from existing methodologies, we propose an innovative unsupervised behavior anomaly detection framework named Multilevel Guided Exploration Network(MGENet), which focuses on exploring high-level feature difference rather than recovering or predicting pixel-level information.  As shown in Figure.~\ref{fig:introduction}(c), MGENet detects motion and appearance anomalies using a Two-level guidance-exploration pattern. Each level is trained with different types of input features and leverages the difference in output features between the Guidance and Exploration Networks to detect anomalies. This design makes it challenging for anomalies to exhibit performance similar to normal samples.

Specifically,  we employ Spatio-temporal Normalizing Flow~\cite{STG-NF} to map normal human-pose data into a latent representation characterized by a Gaussian distribution. This process strategically situates anomalous pose data at the distribution's periphery. Then, guided by Normalizing Flow, the RGB Encoder captures spatio-temporal features,  detecting motion anomalies by analyzing the difference in the output features between these two networks.
Furthermore, the RGB Encoder also guides the unmask Encoder to distill high-level features from specific patches of masked RGB frames, detecting appearance anomalies based on the similarity between the high-level features output by both networks. Additionally, we incorporate a Behavior-Scene matching module, which establishes and stores the relationship between normal behavior and scenes, enabling the detection of scene anomalies. We demonstrate the effectiveness of the method on two publicly available datasets.


\section{Related Work}

\subsection{Video Anomaly Detection}
In recent years, numerous studies have achieved remarkable results based on RGB frames, optical flow, or pose features. 

Some researchers employ reconstruction methods\cite{memory,hsc,sspcab,ssmtl,liu2021hybrid}, for anomaly detection, assessing anomalies based on higher reconstruction errors compared to normal samples.  ~\citet{memory} propose augmenting the autoencoder with a memory module, favoring proximity to normal samples during reconstruction and amplifying errors for anomalies. \citet{hsc}  utilize two autoencoders to reconstruct motion and appearance features. Furthermore, they also design a contrastive learning method to identify scene-related behavioral anomalies, but they only detect within limited scenarios, lacking diversity.
Some researchers use prediction methods\cite{liu2018future,Consistency,aaai21-Consistency,keyframe,ssmtl++,liu2021hybrid}, to detect anomalies.~\citet {liu2018future} propose a future frame prediction approach, which detects anomalies by assessing the discrepancies between predicted images and actual images. Chen et al.~\cite{chen2022comprehensive} find limitations in simple prediction constraints for representing appearance and flow features. They introduce a novel bidirectional architecture with three consistency constraints to regulate the prediction task. ~\citet{keyframe} introduced the task of key frame restoration, encouraging Deep Neural Networks to infer missing frames based on video key frames, thereby restoring the video. 

Furthermore, in recent years, there has been an emergence of utilizing alternative methods for anomaly detection. Hirschorn et al.~\cite{STG-NF} employ Normalizing Flow to map normal data into the latent representation, locating anomalous data at the distribution periphery. Wang et al.~\cite{jigsaw} propose a new pretext task, disrupting both temporal and spatial order and training the model to restore RGB frames. 

\begin{figure*}[t]
    \centering
    \setlength{\abovecaptionskip}{0.cm}
    \begin{center}
        \includegraphics[width=1\linewidth]{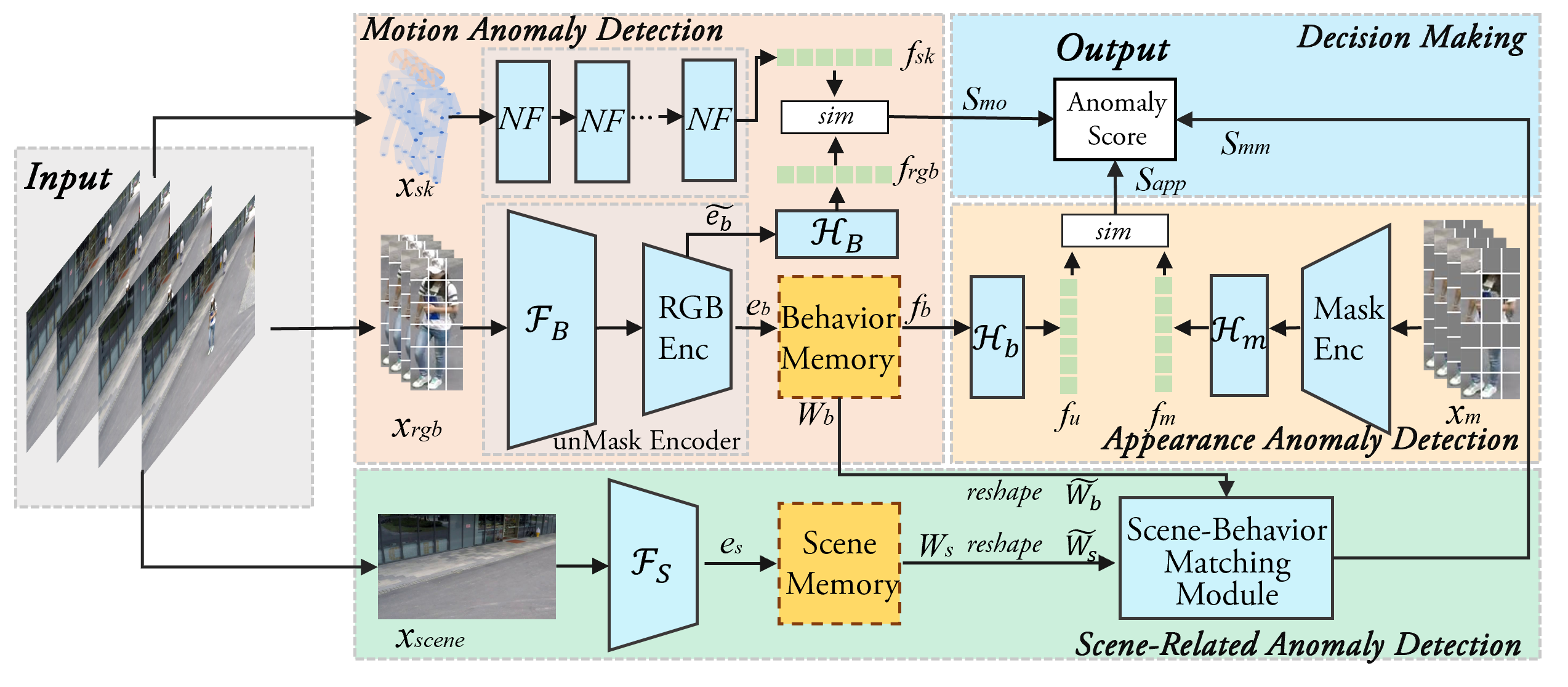}
    \end{center}
    \caption{The overall framework of our method.}
    \label{fig:framwork}
\end{figure*}

\subsection{Student Teacher Network}
Recently, in the field of industrial anomaly detection, some researchers ~\cite{AST} employ knowledge distillation to detect anomalies by utilizing the regression error of student networks on the feature outputs of a high-parameter teacher network. Specifically, the STPM method~\cite{RST} is based on student-teacher feature pyramid matching, with the student and teacher networks being pre-trained as ResNet50 and ResNet18, respectively. In this scenario, the student model, with fewer parameters, closely approximates the teacher's performance on normal data but exhibits notable disparities when encountering unseen anomalous data.

The methods mentioned earlier and our approach's appearance anomaly detection phase share similarities, employing knowledge distillation. However, there's a difference in the motion anomaly detection process, where knowledge distillation is not the primary emphasis. The Normalizing Flow has a lightweight architecture, whereas the RGB Encoder has a more complex structure. Additionally, there are substantial differences in the frameworks and input data types between these two networks.

\subsection{Masked Visual Model}
Masked Visual Modeling~\cite{wei2022mvp} improves visual representation learning by masking image portions. 
Chen et al.~\cite{chen2023cae} propose a pretraining method with two tasks: predicting representations for masked patches and reconstructing masked patches. Zhang et al.~\cite{zhang2022caev2} demonstrate improved performance with only supervised visible patches, omitting the need for masked patches. In the realm of video representation learning, Tong et al.~\cite{tong2022videomae}show that video-masked autoencoders are also data-efficient learners for self-supervised video pre-training. Wang et al.~\cite{wang2023maskedvideo}  present Masked Video Distillation, a succinct two-stage framework, for video representation learning. Inspired by masking tasks, we mask partially video frames and exclusively use visible frames to learn the latent high-level features of the uncovered frames. In this way, since the model has not encountered the appearance of anomalies before, there will be differences in the latent representation of unmasked frames.
\section{Method}
\subsection{Overview}

Figure ~\ref{fig:framwork} illustrates the overall framework. Given a video clip with $T$ consecutive frames, we extract human-centric RGB frames $x_{rgb} \in \mathbb{R}^{T \times H \times W\times C}$ and $V$-joints skeletal pose data $x_{sk} \in \mathbb{R}^{T \times V \times C}$~\cite{fang2022alphapose,xiu2018poseflow}. Meanwhile, according to BEIT~\cite{bao2021beit}, we segment the RGB frames into $P$ patches and mask the rate of $\gamma$ of the patches to obtain $x_{m}$. Then, anomalies are detected through the following four processes.

\textbf{Motion Anomaly Detection:} First, we pre-train Normalizing Flow to project $x_{sk}$ into a latent representation $f_{sk} \in \mathbb{R}^{T \times C_{mo}}$ following the Gaussian distribution. Then, it guides the RGB Encoder and head $\mathcal{H}_B$ using RGB frames to learn spatio-temporal features $f_{rgb}  \in \mathbb{R}^{T \times C_{mo}}$ 

\textbf{Appearance Anomaly Detection:} Behavior Features $e_{b}$ pass through behavior memory to obtain  feature $f_b$, then we use head $\mathcal{H}_{u}$ to map  $f_b$ into $f_{u}\in \mathbb{R}^{T \times  C_{app}} $. Following this, it guides the Mask Encoder using masked RGB frames $x_{m}$ to learn appearance features $f_{m}  \in \mathbb{R}^{T \times C_{app}}$.

\textbf{Scene-related Anomaly Detection:} The scene undergoes feature extraction $\mathcal{F}_S$ to generate scene features, alongside corresponding behavior features, pass through the memory banks to get soft addressing weights $W_b \in \mathbb{R}^{P\times N_b} $ and $W_s \in \mathbb{R}^{1\times N_s}$, respectively. Here, \(N_b\) and \(N_s\) represent the number of slots in the Behavior and Scene Memory, respectively. Then, they pass into the Behavior-Scene Matching Module together to compute the scene-related anomaly score $S_{mm}$.

\textbf{Anomaly Score:} The score is computed by considering the difference between the pose feature $f_{sk}$ and the behavior feature $f_{rgb}$, the similarity of appearance features $f_{u}$ and $f_m$, and the matching score $S_{mm}$.

\subsection{Motion Anomaly Detection}
Skeletal data helps the model capture the essential characteristics of movements or postures\cite{lee2023hierarchically-sk,STGCN}. Recently, Hirschorn et al.\cite{STG-NF} designed the Spatio-temporal Normalizing Flow($\textit{NF}_s$), including $L$ flow modules, which can map the skeletal distribution of normal skeletal data to a standard distribution through a series of invertible transformations, with anomalies typically found at the distribution's periphery.  

Building on the above rationale, firstly, we train Spatio-temporal Normalizing Flow  according to~\cite{STG-NF}, mapping pose data $x_{sk}$ to latent behavior features $f_{sk}$ : 
\begin{align}
    f_{sk} = \textit{NF}_s(x_{sk}).
\end{align}
Secondly, we use it as a pre-trained model to guide the exploration network (RGB Encoder) to generate the latent motion features $f_{rgb}$. The following are the detailed steps:

First, given RGB frames $x_{rgb} \in \mathbb{R}^{T \times H \times W \times  C}$, similar to  cube embedding~\cite{arnab2021vivit,tong2022videomae,dosovitskiy2020vit}, we treat each cube of $2\times 8\times 8$ as one token embedding, and  obtain $t\times h \times w$ 3D tokens,where $t = \frac{T}{2}$,$h =\frac{H}{8}$, $w = \frac{W}{8}$. Then, map each token to the channel dimension. Next, we pre-extract RGB features of these tokens and employ the RGB Encoder, a VIT backbone with joint space-time attention~\cite{dosovitskiy2020vit,tong2022videomae}, to obtain spatio-temporal features $e_b \in \mathbb{R}^{P \times  C_{b}}$, where $P= t \cdot h \cdot w$.

Then, we reshape $e_b$ into $\tilde{e}_b \in \mathbb{R}^{t \times  C_{b} \times h \times w }$ and design the spatial-temporal head $\mathcal{H}_B$, which replaces the $3\times3\times3$ convolution in Spatial-Temporal Excitation~\cite{std} with the decomposed Large kernel Attention, named large Spatial-Temporal Attention(LSTA),  to further capture the spatio-temporal relationships of patches with long-distance temporal dependencies and spatial variations in different frames of human actions.

\begin{figure}[t]
    \centering
    \setlength{\abovecaptionskip}{0.cm}
    \begin{center}
        \includegraphics[width=0.85\linewidth]{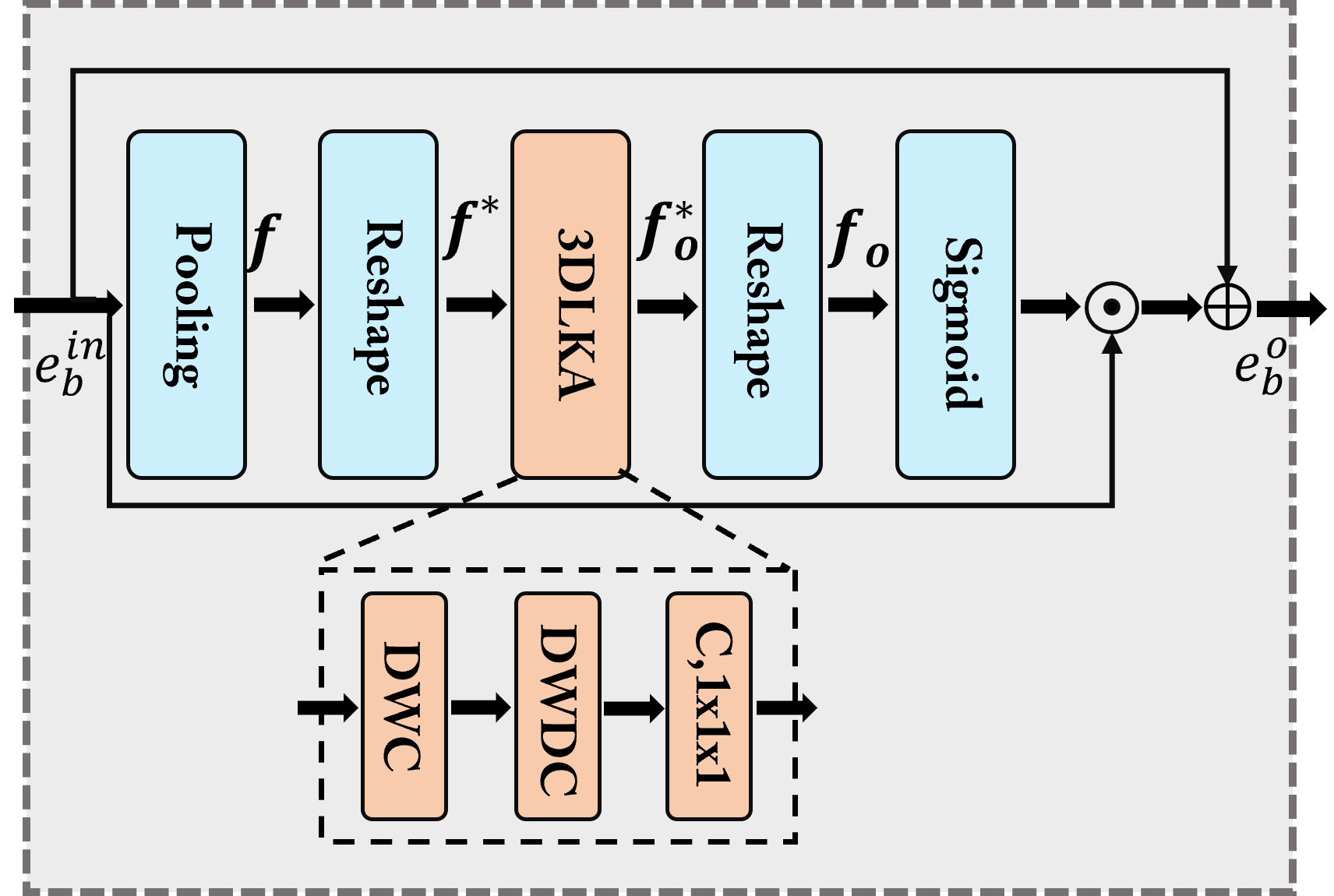}
    \end{center}
    \caption{The framework of LSTA.}
    \label{fig:rvan}
\end{figure}

In detail,spatio-temporal head $\mathcal{H}_B$  consists of $L$ LSTA modules and the MLP layers. As shown in Figure \ref{fig:rvan}, given an input tensor $e^{in}_b \in \mathbb{R}^{t \times c \times h \times w}$, we begin by performing channel-wise averaging, yielding a global spatio-temporal tensor $f \in \mathbb{R}^{ t \times 1 \times h \times w}$. Then, we reshape $f$ into $f^* \in \mathbb{R}^{ 1 \times t  \times h \times w}$ and pass it through the 3DLKA module to get transformed tensor $f^*_o \in \mathbb{R}^{ 1 \times t  \times h \times w}$,which is represented as follows :
\begin{align}
   f^{*}_{o} = \textit{3DLKA}(f^{*})= \textit{CONV}(\textit{DWDC}(\textit{DWC}(f^{*}))),
\label{eq}
\end{align}
where $\textit{DWC}$ denotes  a $\frac{k}{d} \times \frac{k}{d} \times \frac{k}{d} $ deep dilated convolution with dilated $d$ ,$\textit{DWDC}$ denotes a $(2d - 1) \times (2d - 1)\times (2d - 1)$ deep convolution, and $\textit{CONV}$  denotes a $1 \times 1 \times 1$ convolution.
Subsequently, $f^*_o$ is reshaped to $f_o \in \mathbb{R}^{ t \times 1 \times h \times w}$ and passed through a sigmoid activation function to obtain the attention map. Finally, we use this map to guide $e^{o}_b $ for obtaining behavior feature $e^{o}_{b}$:
\begin{align}
    e^{o}_b = e^{in}_b + e^{in}_b \odot sigmoid(f_o),
\label{eq}
\end{align}
where $\odot$ denotes the element-wise product. After passing through the MLP layers, we obtain the motion feature $f_{rgb}$. Finally, we minimize the difference between  $f_{sk}$ and $f_{rgb}$ feature to facilitate the model in learning spatio-temporal pose features of normal patterns.
\begin{align}
     \mathcal{L}_{mo} = ||f_{sk} - f_{rgb} ||^2_2.
\end{align}
In this way, for anomalous samples, achieving similar high-level semantic representation is more challenging due to differences in feature modalities and network architectures. Therefore, we can detect action anomalies based on the difference between the two types of features.

\subsection{Appearance Anomaly Detection}

Beyond motion anomalies, our method considers appearance anomalies, including carrying unidentified objects or using inappropriate vehicles. We extend Masked Image Modeling (MIM) \cite{xie2022simmim} to video anomaly detection, enabling the Mask Encoder to learn normal appearance features guided by the unmasked RGB Encoder. This adaptation tackles challenges faced by the Mask Encoder in capturing high-level features of patches that were not encountered before but now are masked, resulting in noticeable differences from unmasked RGB features.

Specifically, following the approach of BEIT $\textit{MASK}$ ~\cite{bao2021beit}, we mask patches with a ratio $\gamma$, which is set to 50\% and obtain masked RGB frames $x_{m}$. Noted that we used the same mask for the frames within the same video frame, preventing the model from extracting patch features from adjacent frames.
\begin{align}
    \{ x^{1}_{m},x^{2}_{m},...,x^{P/2}_{m}\} =  \textit{MASK}\{ x^{1}_{rgb},x^{2}_{rgb},...,x^{P}_{rgb}\},
\end{align}
where $P$ denotes the number of patches. Next, we also use the cube embedding method described in Section 3.2 to obtain tokens. These tokens pass through the Mask Encoder, which has a structure similar to the RGB Encoder, to learn appearance features. Then, the projection head is employed to obtain latent appearance features, denoted as $f_{m}$. Simultaneously, the appearance features $f_{b}$ also undergo the projection head to obtain latent appearance features $f_{u}$ with the same size as $f_{b}$.

During the training process, the RGB Encoder keeps fixed, while the Mask Encoder distills the high-level feature representation of the RGB Encoder under the condition of having only partially visible patches. The final loss function can be expressed as:
\begin{align}
     \mathcal{L}_{app} = 1 - \frac{{f_{m}} \cdot {f_{u}}}{\|{f_{m}}\| \cdot \|{f_{u}}\|}.
\end{align}
\subsection{Scene-Releted Anomaly Detection}
Formally, we postulate that unobserved behaviors within a scene should be categorized as anomalies. To detect these anomalies, we design the Scene-Behavior Matching Module to capture the relationships between normal patterns of scenes and behaviors. As a result, scene-related behavior anomalies exhibit weaker matching with the learned features, leading to higher anomaly scores.

\begin{figure*}[t]
    \centering
    \setlength{\abovecaptionskip}{0.cm}
    \begin{center}
        \includegraphics[width=0.96\linewidth]{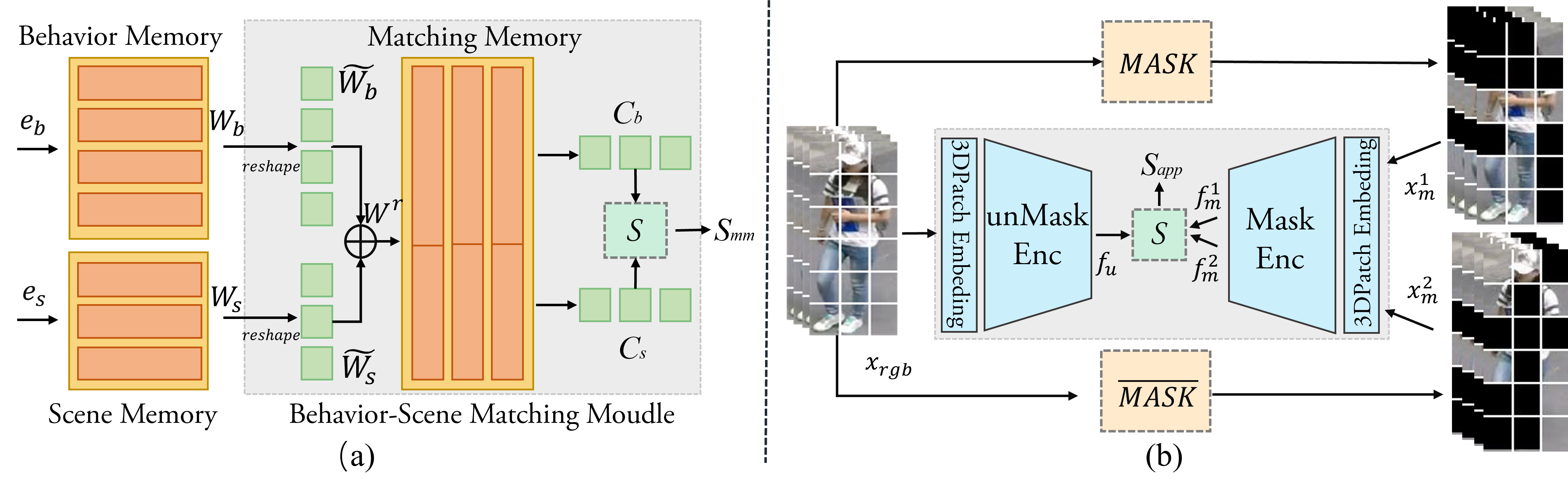}
    \end{center}
    \caption{Calculation process of (a) scene-related anomaly score and (b) appearance anomaly score. Here, S represents similarity calculation,$\textit{MASK}$ and $\overline{\textit{MASK}}$ represent mutually opposite masks. Note that in figure(b), the two sets of masked images are sequentially processed through the Mask Encoder}
    \label{fig:test}
\end{figure*}

As illustrated in Figure.~\ref{fig:test}(a), the Behavior-Scene Matching Memory(BSMM), similar to the Behavior Memory and Scene Memory, is a read-write memory with a similar structure~\cite{memory}. However, the key difference is that the Behavior-Scene Matching Memory stores the representation of the addressing weights in the Behavior Memory and Scene Memory for the behavioral features and their corresponding scene features of all normal data. Below is an introduction to its update and read processes.

First, the scene image passes through the feature extractor, generating scene features \(\mathbf{e}^s \in \mathbb{R}^{C_s} \). Then, behavior features $\mathbf{e}^b  \in \mathbb{R}^{P \times C_b}$ and $\mathbf{e}^s$  query the Behavior Memory and Scene Memory, respectively, and then contribute to calculating the similarity weights $W_b \in \mathbb{R}^{P \times N_b}$ and $W_s  \in \mathbb{R}^{1 \times N_s}$, where $N_b$ and $N_s$ denote as the number of the slots in behavior memory and scene memory, respectively. For the $i$-th slot in behavior memory, denoted as  $\mathbf{m}^{b}_{i} \in \mathbf{M}^{b}$, we can calculate the addressing weights between it and the $p$-th query $\mathbf{e}^{b}_{p}$ item as follows:
\begin{align}
    w_{i,p}^{b}=\frac{\exp \left(d\left(\mathbf{e}^{b}_{p}, \mathbf{m}_{i}^{b}\right)\right)}{\sum_{j=1}^{N_b} \exp \left(d\left(\mathbf{e}^{b}_{p}, \mathbf{m}_{j}^{b}\right)\right)}
    \label{eq:weight},
\end{align}
where $\operatorname{d}(\boldsymbol{e^*}, \boldsymbol{m^*}) $ denotes cosine similarity. The computation method for ${w_{i}^{s}} \in {W^{s}}$ is the same as that for ${w_{i,p}^{b}} $.

\textbf{Update:} We reshape $W^{b} $ and $W^{s}$ into one-dimensional vectors $\tilde{W}^{b} \in \mathbb{R}^{L_b} $ and $\tilde{W}^{s} \in \mathbb{R}^{ L_s}$ ,where $L_b = P \times C_b$ and $L_s = 1 \times C_s $. Then, concatenate $\tilde{W}^{b}$ and $\tilde{W}^{s}$ along the channel:
\begin{align}
   \mathbf{W}^{r} = [\tilde{W}^{b},\tilde{W}^{s}].
\end{align}
Next, similar to ~\cite{memory,hsc}, for each items $\mathbf{m}^{r}_{i} \in \mathbf{M}^{r}$ in the Behavior-Scene Memory, we update as following:
\begin{align}
   \mathbf{m}^{r}_i \leftarrow f\left(\mathbf{m}^{r}_i+\sum_{v \in U^i} v^{ p, i} \mathbf{W}^{r}_v\right),
\end{align}
where $f(\cdot)$ is the $L_2$ norm. $U^i$ represents the set of indices for the corresponding queries for the $i$-th item in the memory. $v^{p,i}$  represents matching probability between memory items and queries, similar to equation (\ref{eq:weight}). It is worth noting that the aforementioned update operation occurs only in the final round and keeps the parameters of the Behavior Memory and Scene Memory unchanged.

\textbf{Read:} 
Different from the previous method\cite{memory}, we compute the matching weights $c_{i}^{b} \in C^{b} $ between $W_b$ and the first $N_b$ channels of the Matching Memory $\mathbf{M}^{r}$:
\begin{align}
    c_{i}^{b}=\frac{\exp \left(d\left(W^{b}, \mathbf{m}^{r}_{i,:L_b}\right)\right)}{\sum_{j=1}^N \exp \left(d\left(W^{b}, \mathbf{m}^{r}_{j,:L_b}\right)\right)}.
    \label{eq2}
\end{align}
Similarly, we can use the above method with the last $N_s$ channels of $\mathbf{M}^{r}$ to calculate $C^{s}$. 

In this way, $W^b$ and $W^s$ serve as vector representations of historical behaviors and scenes. Their combination stored in $\mathbf{M}^{r}$ forms a pattern of the behavior-scene pattern. During the test phase, $W^b$ and $W^s$ act as query terms, individually computed with the behavioral and scene representations of each item $\mathbf{m}^{r}_i$ in $\mathbf{M}^{r}$ to derive $c^b_i$ and $c^s_i$.  If the difference between them is significant, it indicates a mismatch between the behavior and the current scene. Finally, anomalies are measured by considering all patterns stored in the Matching Memory.



\subsection{Loss Function and Anomaly Score}
\textbf{Loss Function:} The training loss includes the regression loss $\mathcal{L}_{mo}$, and the distillation loss  $\mathcal{L}_{app}$. Additionally, to allocate similar queries to the same item, the objective is to reduce the number of items and the overall memory size according to ~\cite{memory}, there is the feature separateness Loss defined with a margin of $\epsilon$ as follows:
\begin{align}
    \mathcal{L}_{\text{sep}} \!= \!\sum_p^P \left[\left\|\mathbf{W}_p^r \!-\! \mathbf{m}^{st}\right\|_2 \!-\! \left\|\mathbf{W}_p^r \!-\! \mathbf{m}^{nd}\right\|_2 \!+\! \epsilon\right]_{+},
\end{align}
where  \(P\) represents the number of queries, and \(\mathbf{m}^{st}\) and \(\mathbf{m}^{nd}\) represent the first and second nearest items for the query \(\mathbf{W}_p^r\). Thus, for the three memories, the separateness loss is denoted as \(\mathcal{L}^{b}_{sep}\), \(\mathcal{L}^{s}_{sep}\), and \(\mathcal{L}^{r}_{sep}\) respectively. In summary, the overall loss function is expressed as:
\begin{align}
    \mathcal{L} =  \mathcal{L}_{mo} + \alpha \mathcal{L}_{app} + \beta(\mathcal{L}^{b}_{sep}+\mathcal{L}^{s}_{sep}+\mathcal{L}^{r}_{sep}),
\end{align}
where $\alpha$ and $\beta$  are balancing hyper-parameters. 

\textbf{Anomaly Score:}
Measuring the anomaly scores involves three components: motion anomaly score, appearance anomaly score, and scene-related anomaly score. In the first level of our framework, we obtain high-level skeleton feature $f_{sk}$ and behavior feature $f_{rgb}$. Due to the distinct structures and input data of the two modules, When encountering previously unseen anomalous behaviors, there is a substantial difference between them, We can utilize this difference as the Motion Anomaly Score:
\begin{align}
     \mathcal{S}_{mo} = ||f_{sk} - f_{rgb} ||.
\end{align}
Furthermore, as shown in Figure.~\ref{fig:test}(b), given the adoption of a 50\% masking approach, there are two sets of mutually exclusive masks $\textit{MASK}$ and $\overline{\textit{MASK}}$ to ensure complete coverage for all patches.Therefore, the calculation method for appearance anomaly scores  $\mathcal{S}_{app}$ is:
\begin{align}
     \mathcal{S}_{app} = \frac{1}{2} sim(f_{u} , f^{1}_{m} ) +  \frac{1}{2} sim(f_{u} , f^{2}_{m}),
\end{align}
where $sim(f_u,f_{*}) = 1 -f_u\cdot f_{*}/(||f_u||\cdot ||f_{*} ||)$. Next, we can determine scene-related anomalies based on the difference in matching weights $C_b$ and $C_s$ between behavior and scene:
\begin{align}
    S_{mm} = ||C_{b} - C_{s}||.
\end{align}
Taking all the above into consideration, the anomaly score for behavior can be expressed as:
\begin{align}
    Score =   S_{mo} + \lambda_{app} S_{app} + \lambda_{mm}  S_{mm},
\end{align}
where $\lambda_{app}$ and $\lambda_{mm}$  are balancing hyperparameters. Finally, the scores are normalized to the range of 0-1 using min-max scaling. We employ the overlap sampling method, where the score of each video clip in a segment is used as the frame-level score for the intermediate frames.

\section{Conclusion}

We design a novel framework, the multilevel guidance-exploration framework, which combines RGB and skeletal features to detect various anomalies. First, guided by the Normalizing Flow, the RGB Encoder learns high-level motion features from unmasked frames. Simultaneously, guided by the RGB Encoder, the Mask Encoder distills appearance features using masked RGB frames. Then, motion and appearance anomalies are detected based on the similarity between high-level representations. Additionally, we also propose the Scene-Behavior Matching Module to explore the relationship between normal patterns of behaviors and scenes, enabling the detection of behavior anomalies related to scenes. Our approach achieves the best performance on the ShanghaiTech and UBnormal datasets.

{
    \small
    \bibliographystyle{ieeenat_fullname}
    \bibliography{main}
}

\end{document}